\documentclass[MLA,Times2COL]{WileyNJDv5}
\usepackage{booktabs,longtable}
\usepackage{graphicx}
\usepackage{xcolor}

\articletype{LETTER}%

\newcommand{\revision}[1]{{\color{black}#1}}
\newcommand{\revisionfloat}{\color{black}}

\received{00}
\revised{00}
\accepted{00}

\begin{document}

\title{Semantic RGB--Depth Based Surgical Skill Assessment in Microscopic Stereo Videos}

\author[1]{Jecia Z. Y. Mao}
\author[1]{Sue M. Cho}
\author[2]{Francis X. Creighton}
\author[2]{Deepa Galaiya}
\author[1]{Russell H. Taylor}
\author[1]{Manish Sahu}

\titlemark{Semantic RGB--Depth Based Surgical Skill Assessment in Microscopic Stereo Videos}

\address[1]{\orgdiv{Laboratory of Computational Sensing and Robotics},
\orgname{Johns Hopkins University},
\orgaddress{\state{MD}, \country{USA}}}

\address[2]{\orgdiv{Department of Otolaryngology--Head and Neck Surgery},
\orgname{Johns Hopkins University School of Medicine},
\orgaddress{\state{MD}, \country{USA}}}

\corres{Jecia Z. Mao, Johns Hopkins University, 3400 N Charles St, Baltimore, MD 21218, USA.
\email{zmao16@jh.edu}}

\abstract[Abstract]{
Objective assessment of microsurgical technical skill is essential for competency-based training and quality assurance, yet existing video-based approaches predominantly rely on RGB images and therefore overlook the 3D spatial relationships that characterize instrument--anatomy interactions. Although stereo operating microscopes provide complementary depth information, conventional stereo matching algorithms frequently produce sparse and unreliable depth estimates under high-magnification imaging conditions, limiting their use for automated skill assessment. This work presents a semantic RGB--Depth approach for video-based surgical skill assessment from microscopic stereo videos. A regression-based depth fusion method is first introduced to combine sparse metric stereo depth with dense monocular depth estimation to generate a dense geometrical representation of the surgical scene. The reconstructed depth representation is then integrated with semantically decomposed RGB streams corresponding to individual surgical instruments and surrounding anatomy, which are jointly encoded through a hierarchical attention architecture to capture discriminative patterns of instrument use and instrument–anatomy interaction across surgeons at different training levels. The proposed framework was evaluated on 33 \emph{ex vivo} transoral microlaryngeal procedures performed by six surgeons \revision{comprising attending surgeons and surgical residents using leave-one-surgeon-out cross validation}. Incorporating depth representation improved prediction performance over RGB-only representations, and the proposed semantic RGB--Depth model achieved an F1 score of \revision{0.938}, outperforming semantic RGB \revision{(0.696)} and semantic depth \revision{(0.929)} models. These results suggest that geometric information can improve automated surgical skill assessment from microscopic stereo videos. \revision{The learned spatial, temporal, and semantic attention patterns further allow qualitative examination of scene components that contribute to skill assessment.}
}


\maketitle

\renewcommand\thefootnote{\fnsymbol{footnote}}
\setcounter{footnote}{1}

\section{Introduction}
\label{sec:Intro}

Microsurgery requires precise instrument manipulation under high-magnification visualization within confined anatomical environments. Transoral microlaryngeal surgery exemplifies these challenges, requiring surgeons to manipulate delicate vocal fold tissues through a narrow laryngoscope using long microsurgical instruments. Successful performance depends on precise hand-eye coordination, controlled bimanual manipulation, efficient instrument motion, and appropriate instrument--tissue interactions. Even small technical errors can compromise surgical outcomes and result in irreversible injury to critical anatomical structures. Consequently, acquiring microsurgical proficiency requires extensive training, and objective assessment of technical skill has become an important component of competency-based surgical education.

Conventional assessment of microsurgical skill relies primarily on expert observation~\cite{martin1997objective,niitsu2013Objective}. However, they are inherently subjective, time-consuming, and susceptible to inter-rater variability \cite{reznick2006teaching}. 
To address these limitations, automated surgical skill assessment has emerged as an active area of research \cite{zia2018automated,funke2019video,kelly2020bidirectional,wang2018deep,lavanchy2021automation,liu2021towards,li2022surgical,anastasiou2023keep,ding2023sedskill,wang2020towards,zia2018video}. Existing approaches can generally be categorized according to their input modality into instrument kinematics-based~\cite{wang2018deep,anastasiou2023keep}, video-based~\cite{funke2019video,wang2020towards,kelly2020bidirectional,liu2021towards,lavanchy2021automation,li2022surgical,hira2022video,ding2023sedskill,wan2024spatiotemporal}, and hybrid methods~\cite{liu2021towards,wang2020towards,anastasiou2023keep}. Kinematics-based approaches require dedicated tracking hardware or robotic systems that directly record instrument motion, limiting their applicability to robot-assisted procedures. In contrast, video-based approaches operate directly on surgical recordings, making them applicable to both robotic and conventional microsurgery without requiring specialized instrumentation or modifications to the surgical workflow. Given the widespread availability of surgical video archives, video-based methods also provide a scalable framework for retrospective analysis and objective assessment across diverse clinical settings.

Recent advances in deep learning have substantially improved video-based surgical skill assessment. Existing methods can generally be divided into two categories \cite{wan2024spatiotemporal}. End-to-end approaches learn a direct mapping from RGB video sequences to surgical skill scores by combining convolutional neural networks with recurrent or temporal architectures to capture spatiotemporal representations \cite{zia2018automated}. Alternatively, multi-stage methods first extract interpretable intermediate information, such as instrument trajectories, surgical gestures, or surgical phases, before modeling their temporal dynamics for skill prediction \cite{lavanchy2021automation,hira2022video}. 
More recently, semantic feature aggregation has demonstrated that explicitly modeling different surgical scene components can improve both prediction accuracy and interpretability~\cite{liu2021towards,li2022surgical}. Nevertheless, nearly all existing video-based approaches rely exclusively on monocular RGB images and therefore characterize surgical performance only within the two-dimensional image plane. Consequently, critical geometric information describing instrument positioning, motion, and spatial relationships with surrounding anatomy is largely ignored.


Dense 3D scene understanding has the potential to overcome this limitation by providing quantitative geometric information that complements visual appearance. However, obtaining reliable dense depth measurements from microscopic stereo video remains challenging. Conventional stereo matching algorithms often generate sparse or noisy disparity estimates when applied to high-magnification surgical microscopes~\cite{wisotzky2025intraoperative,ha2024target}.
As a result, despite the routine availability of stereo microscopes, current video-based assessment methods have largely remained restricted to two-dimensional image representations. \cite{cho2024tremor,wan2024spatiotemporal}



In this work, we address this gap by introducing a semantic RGB--Depth approach that leverages dense 3D scene representation for surgical skill assessment from microscopic stereo videos. First, we propose a regression-based depth fusion method that combines sparse stereo depth with dense monocular depth estimation to reconstruct dense, depth maps of the surgical scene. Rather than treating the surgical video as a single visual input, we further decompose each frame into semantically meaningful regions corresponding to individual surgical instruments and surrounding anatomy. RGB appearance and reconstructed depth are independently encoded within each semantic stream before being integrated through hierarchical temporal and stream-wise attention mechanisms. 
\revision{This design allows the model to combine visual appearance and scene geometry while learning discriminative patterns of instrument use and instrument--anatomy interaction across surgeons at different training levels.}

The proposed approach is evaluated on a dataset comprising 33 \textit{ex vivo} transoral microlaryngeal procedures performed on cadaveric pig larynges by six surgeons with \revision{two distinct skill levels using a leave-one-surgeon-out cross validation}. Experimental results show that dense depth provides complementary information to RGB appearance, while the proposed semantic RGB--Depth approach achieves the strongest overall performance. 
The learned attention weights are also analyzed across semantic streams, time, and image regions to examine which components of the surgical scene receive greater model attention during prediction.

The main contributions of this work are:
\begin{itemize}
    \item A \textbf{regression-based dense depth estimation method} that combines sparse metric stereo reconstruction with dense monocular depth estimation to recover metrically consistent depth maps from high-magnification microscopic stereo videos.
    \item A \textbf{semantic RGB--Depth model} that jointly incorporates visual appearance and reconstructed three-dimensional scene geometry through parallel semantic instruments and anatomy streams with hierarchical attention for video-based surgical skill assessment.
    \item An \textbf{evaluation on a microlaryngeal surgery dataset}, demonstrating that incorporating depth-derived geometric information improves automated surgical skill assessment from microscopic stereo videos.
\end{itemize}


\begin{figure*}[t]
    \centering
    \includegraphics[width=0.8\linewidth]{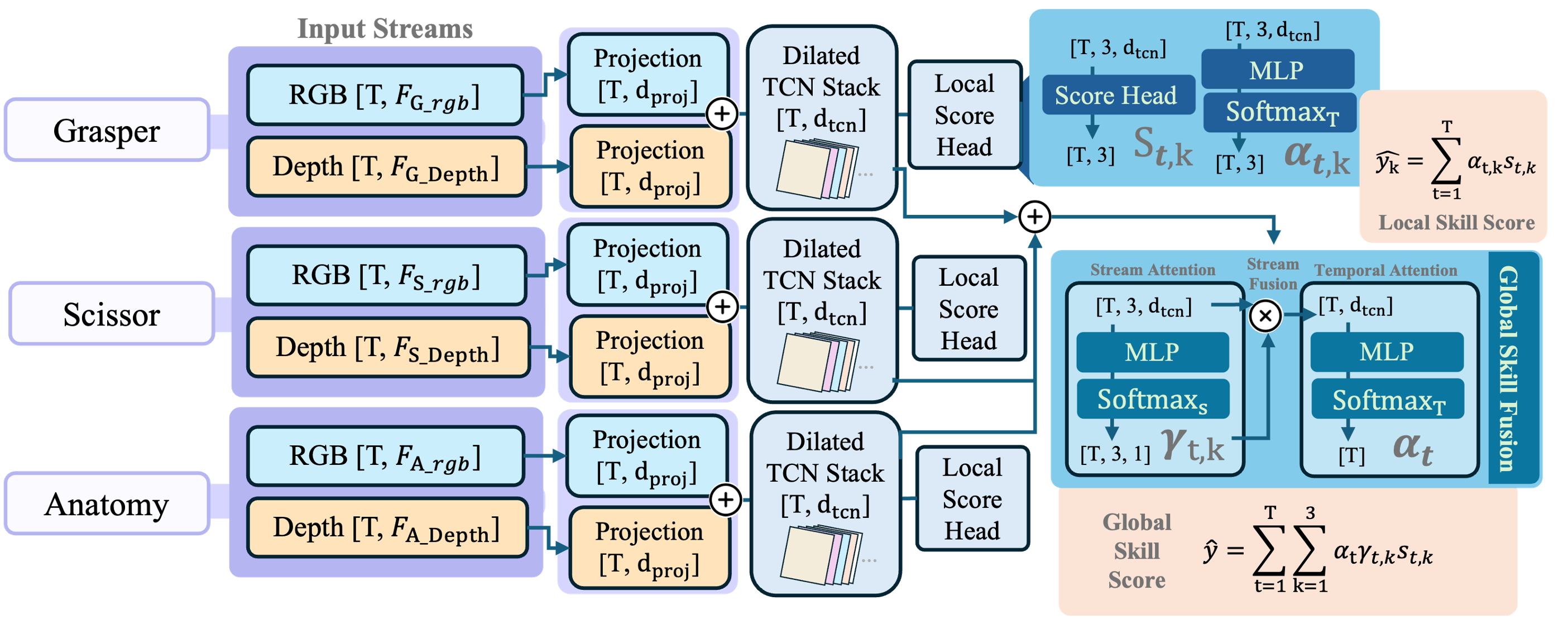}
\caption{
Architecture of the proposed semantic-stream framework. Each trial is decomposed into grasper, scissors, and anatomy streams, indexed by $k\in\mathcal{K}$. RGB and depth features are fused within each stream and encoded by an independent dilated TCN to obtain $\mathbf{h}_{t,k}$. A local score head produces frame-level skill score $s_{t,k}$, while stream-specific temporal attention $\alpha_{t,k}$ forms the auxiliary prediction
$\hat{y}_{k}=\sum_{t=1}^{T}\alpha_{t,k}s_{t,k}$.
For global prediction, frame-wise stream attention $\gamma_{t,k}$ fuses the semantic representations into
$\bar{\mathbf{h}}_{t}=\sum_{k\in\mathcal{K}}\gamma_{t,k}\mathbf{h}_{t,k}$.
Global temporal attention $\alpha_t$ then identifies informative frames, and the final trial-level prediction is
$\hat{y}=\sum_{t=1}^{T}\sum_{k\in\mathcal{K}}\alpha_t\gamma_{t,k}s_{t,k}$.
Auxiliary stream-level auxiliary loss encourages each semantic branch to retain independently discriminative skill information.
}
\label{fig:model_overview}
\end{figure*}

\section{Methodology}
\label{sec:method}


Fig.~\ref{fig:model_overview} provides an overview of the proposed framework. Given a rectified stereo microscopic video, the objective is to \revision{combine RGB appearance and depth-derived geometric information from individual surgical instruments and the surrounding anatomy} for surgical skill assessment. To achieve this, the framework first reconstructs a dense metric representation of the surgical scene by combining stereo and monocular depth estimation. The reconstructed scene is then decomposed into semantically meaningful regions corresponding to the surgical instruments and surrounding anatomy. Finally, RGB appearance, geometric depth, and instrument motion are jointly encoded through a hierarchical attention architecture to predict surgical skill.

\revision{
Unlike conventional video-based methods that operate solely on monocular RGB images, the proposed approach incorporates depth-derived geometric information, enabling the model to explicitly capture discriminative instrument appearance, scene geometry, and instrument-tissue interactions.
}

\subsection{Dense Metric Foreground Reconstruction}
\label{sub:foreground_reconstruction}

\paragraph*{Instrument Segmentation and Mask Propagation}
\label{sub:segmentation}

The first frame of each video is initialized using prompt-based segmentation.
Sparse prompts are provided to the Segment Anything Model
(SAM)~\cite{kirillov2023segment} to segment the grasper, scissors, and visible
foreground anatomy. These initial masks establish the semantic identities of
the three scene components used by the downstream model.

The first-frame masks are propagated through the complete sequence using
CUTIE~\cite{cheng2024putting}, producing frame-wise semantic masks
$\{M_{t,k}\}_{t=1}^{T}$ for each stream $k\in\mathcal{K}$. Here,
$M_{t,k}\in\{0,1\}^{H\times W}$ identifies the image region associated with
the grasper, scissors, or anatomy at frame $t$. The propagated masks are used
to isolate the corresponding RGB and depth information for semantic feature
extraction.
\begin{figure*}[t]
\centering
\includegraphics[width=0.9\textwidth]{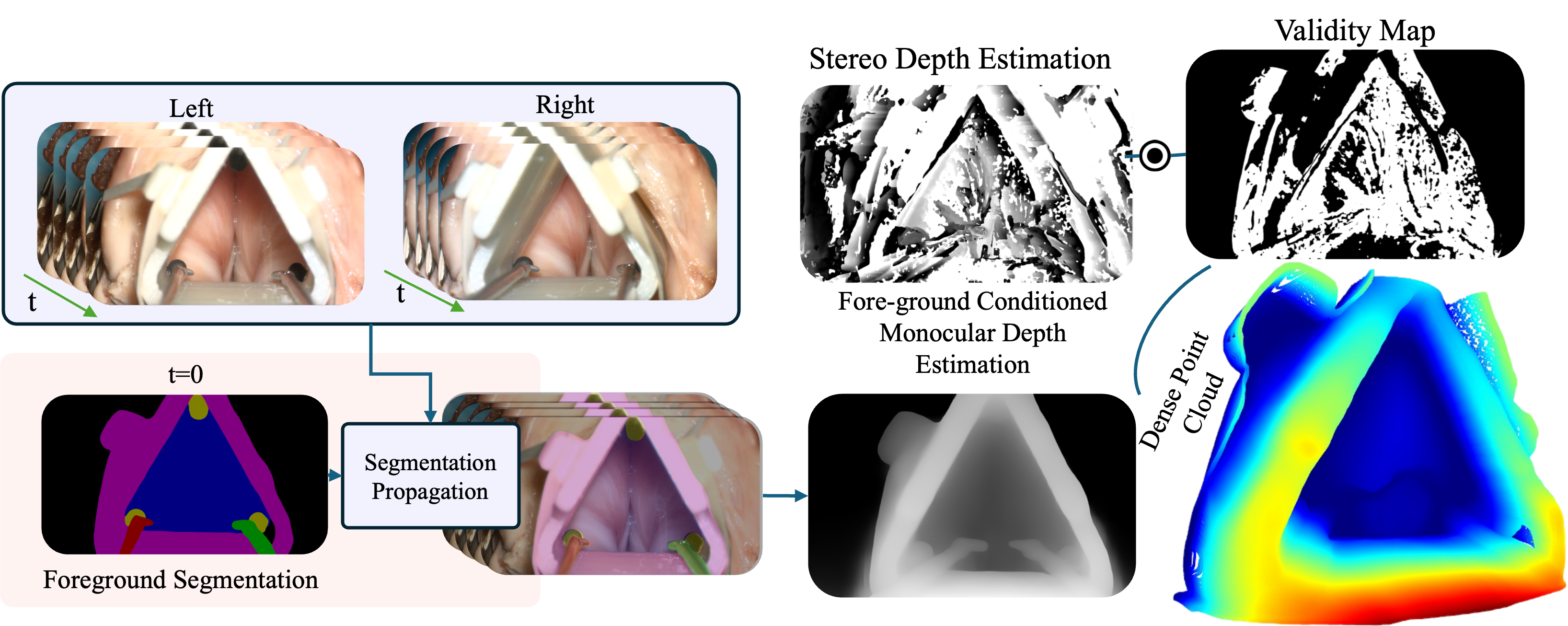}
\caption{Overview of the proposed depth-guided semantic feature extraction
pipeline. The first frame of each microscopic stereo video is annotated using
prompt-based segmentation, and the resulting instrument and foreground masks
are propagated throughout the sequence. Stereo depth provides sparse metric
measurements, while monocular depth provides dense relative structure. The two
estimates are aligned within the foreground region to recover dense metric
depth. Semantic RGB and depth features are then extracted separately for the
grasper, scissors, and anatomy streams and used for temporal skill assessment.}
\label{fig:system_overview}
\end{figure*}

\paragraph*{Dense Metric Depth Estimation}
\label{sub:depth_fusion}

\revision{
For each rectified stereo pair $(I_t^L,I_t^R)$, we estimate an initial metric
stereo depth map $Z_t^{s,\ell}$ for view $\ell\in\{L,R\}$ using the OpenCV
block-matching stereo implementation~\cite{opencv_stereobm}. From the resulting
disparity map, we construct a binary validity mask $C_t^\ell$ by retaining
pixels with positive disparity estimates and rejecting zero or invalid
disparities. Valid stereo measurements are subsequently restricted to the
segmented operative region during metric alignment, such that only
correspondences within the region of interest are retained. Because stereo
matching can be incomplete or noisy in microscopic surgical scenes, these
measurements are treated as sparse metric depth references rather than a dense
reconstruction. In parallel, a dense relative depth map $Z_t^{m,\ell}$ is
estimated using Depth Anything V2~\cite{yang2024depthanythingv2}.

Reliable foreground pixels used for metric alignment are defined as
\begin{equation}
\Omega_t^\ell =
\left\{
\mathbf{p}
\;\middle|\;
M_t^\ell(\mathbf{p})=1,\;
C_t^\ell(\mathbf{p})=1
\right\},
\end{equation}
where $\mathbf{p}=(u,v)$ denotes an image pixel, $M_t^\ell$ denotes the
segmented operative-region mask, and $C_t^\ell$ denotes the binary stereo
validity mask.
}

The dense metric depth map is then obtained as
\begin{equation}
\hat{Z}_t^\ell(\mathbf{p})
=
\alpha_t^\ell Z_t^{m,\ell}(\mathbf{p})+\beta_t^\ell.
\end{equation}
This preserves the metric scale provided by stereo while using monocular
depth to fill regions where stereo matching is unreliable or unavailable.

\subsection{Semantic-Stream Skill Representation}
\label{sub:appearance_controlled_skill}

Rather than organizing the representation only by feature modality, we
decompose each frame into three semantically meaningful scene components:
\begin{equation}
\mathcal{K}
=
\{
\mathrm{grasper},
\mathrm{scissors},
\mathrm{anatomy}
\}.
\end{equation}
Each stream contains the visual appearance and geometric information
associated with its corresponding operative element. This organization allows
the learned attention and score patterns to be interpreted according to
instrument role and anatomical context.

Let $\mathbf{x}^{r}_{t,k}$ and $\mathbf{x}^{d}_{t,k}$ denote the RGB and
depth features at frame $t$ for semantic stream $k$. Because the two
modalities have different dimensions and statistical properties, they are
projected into compatible latent spaces:
\begin{equation}
\mathbf{r}_{t,k}
=
\phi_{r,k}\left(\mathbf{x}^{r}_{t,k}\right),
\qquad
\mathbf{d}_{t,k}
=
\phi_{d,k}\left(\mathbf{x}^{d}_{t,k}\right),
\qquad
k\in\mathcal{K}.
\end{equation}

\paragraph*{Semantic-Stream Fusion}
\label{sub:semantic_stream_fusion}

The projected RGB and depth features are fused independently within each
semantic stream:
\begin{equation}
\mathbf{u}_{t,k}
=
\phi_{f,k}
\left(
\left[
\mathbf{r}_{t,k};
\mathbf{d}_{t,k}
\right]
\right),
\qquad
k\in\mathcal{K}.
\end{equation}

The frame-level representations of stream $k$ are collected into
\begin{equation}
\mathbf{U}_{k}
=
\left[
\mathbf{u}_{1,k},
\ldots,
\mathbf{u}_{T,k}
\right].
\end{equation}

\paragraph*{Per-Stream Temporal Encoding}
\label{sub:per_stream_temporal_encoding}

Surgical skill is expressed through temporally evolving patterns of
instrument control and tool--anatomy interaction. Each fused semantic stream
is therefore processed by an independent dilated temporal convolutional
network:
\begin{equation}
\mathbf{h}_{t,k}
=
f_{\mathrm{TCN},k}
\left(
\mathbf{u}_{1,k},
\ldots,
\mathbf{u}_{T,k}
\right)_{t},
\end{equation}
where
$\mathbf{h}_{t,k}\in\mathbb{R}^{d_{\mathrm{tcn}}}$
is the temporally encoded representation of stream $k$ at frame $t$.
\revision{
Using a dedicated temporal convolutional network (TCN) for each semantic
stream allows the grasper, scissors, and anatomy branches to learn distinct
temporal dynamics before global fusion.
}

\paragraph*{Local Skill Score}
\label{sub:local_skill_prediction}

\revision{
A stream-specific multilayer perceptron (MLP) score head is applied to each
temporally encoded feature:
}
\begin{equation}
s_{t,k}
=
\mathrm{MLP}_{s,k}
\left(
\mathbf{h}_{t,k}
\right),
\end{equation}
where $s_{t,k}$ denotes the local skill score produced by stream $k$ at
frame $t$.

Each stream also learns an independent temporal-attention distribution:
\begin{equation}
\alpha_{t,k}
=
\frac{
\exp\left(
\mathrm{MLP}_{\alpha,k}
\left(
\mathbf{h}_{t,k}
\right)
\right)
}{
\sum_{\tau=1}^{T}
\exp\left(
\mathrm{MLP}_{\alpha,k}
\left(
\mathbf{h}_{\tau,k}
\right)
\right)
}.
\label{eq:local_temporal_attention}
\end{equation}
The normalization is performed over time within each stream, such that
\begin{equation}
\sum_{t=1}^{T}\alpha_{t,k}=1,
\qquad
\forall k\in\mathcal{K}.
\end{equation}
The local temporal attention $\alpha_{t,k}$ identifies the frames that are
most informative within stream $k$ and is used to form the auxiliary
stream-level prediction:
\begin{equation}
\hat{y}_{k}
=
\sum_{t=1}^{T}
\alpha_{t,k}s_{t,k}.
\label{eq:local_stream_prediction}
\end{equation}

\paragraph*{Frame-Wise Semantic Stream Attention}
\label{sub:stream_attention}

The relative importance of the grasper, scissors, and anatomy streams may
vary throughout a trial. Stream attention is therefore computed directly
from the temporally encoded features:
\begin{equation}
\gamma_{t,k}
=
\frac{
\exp\left(
\mathrm{MLP}_{\gamma}
\left(
\mathbf{h}_{t,k}
\right)
\right)
}{
\sum_{j\in\mathcal{K}}
\exp\left(
\mathrm{MLP}_{\gamma}
\left(
\mathbf{h}_{t,j}
\right)
\right)
}.
\label{eq:stream_attention}
\end{equation}
The normalization is performed across semantic streams at each frame:
\begin{equation}
\sum_{k\in\mathcal{K}}\gamma_{t,k}=1,
\qquad
\forall t.
\end{equation}
Thus, $\gamma_{t,k}$ represents the relative emphasis placed on stream $k$
at frame $t$. In contrast, $\alpha_{t,k}$ is normalized over time and
represents the temporal importance of frame $t$ within stream $k$.

\paragraph*{Global Stream Fusion and Temporal Attention}
\label{sub:global_local_fusion}

The semantic representations are fused at each frame using the learned
stream-attention weights:
\begin{equation}
\bar{\mathbf{h}}_{t}
=
\sum_{k\in\mathcal{K}}
\gamma_{t,k}\mathbf{h}_{t,k}.
\label{eq:semantic_stream_fusion}
\end{equation}
This produces the globally fused temporal sequence
\begin{equation}
\bar{\mathbf{H}}
=
\left[
\bar{\mathbf{h}}_{1},
\ldots,
\bar{\mathbf{h}}_{T}
\right].
\end{equation}

A separate global temporal-attention network is then applied to the fused
sequence:
\begin{equation}
e_{t}^{\mathrm{global}}
=
h_{\alpha}
\left(
\bar{\mathbf{h}}_{t}
\right).
\end{equation}
The global temporal logits are normalized across the complete trial:
\begin{equation}
\alpha_{t}
=
\frac{
\exp\left(e_{t}^{\mathrm{global}}\right)
}{
\sum_{\tau=1}^{T}
\exp\left(e_{\tau}^{\mathrm{global}}\right)
}.
\label{eq:global_temporal_attention}
\end{equation}
Therefore,
\begin{equation}
\sum_{t=1}^{T}\alpha_t=1.
\end{equation}
The global temporal attention $\alpha_t$ measures the importance of frame
$t$ after information from the grasper, scissors, and anatomy streams has
already been combined. It is learned independently from the local
stream-specific attention $\alpha_{t,k}$.

\paragraph*{Trial-Level Skill Prediction}
\label{sub:trial_skill_prediction}

The final trial-level prediction combines the local skill score with the
global temporal and frame-wise stream-attention weights:
\begin{equation}
\hat{y}
=
\sum_{t=1}^{T}
\sum_{k\in\mathcal{K}}
\alpha_t
\gamma_{t,k}
s_{t,k}.
\label{eq:trial_skill_prediction}
\end{equation}
Here, $s_{t,k}$ denotes the skill score produced by stream $k$ at frame
$t$, $\gamma_{t,k}$ determines the relative importance of that stream within
the frame, and $\alpha_t$ determines the importance of the fused frame within
the complete trial. The effective contribution of stream $k$ at frame $t$ is
therefore
\begin{equation}
c_{t,k}
=
\alpha_t\gamma_{t,k}s_{t,k}.
\end{equation}

Importantly, the local temporal attention $\alpha_{t,k}$ is not multiplied
into the final global prediction. Instead, it is used to construct the
auxiliary prediction $\hat{y}_k$ for each individual stream, whereas the
final prediction uses the global temporal attention $\alpha_t$ learned from
the stream-fused representation.

\paragraph*{Training Objective}
\label{sub:training_objective}

\revision{
The global prediction is supervised using the mean-squared error (MSE):
}
\begin{equation}
\mathcal{L}_{\mathrm{fusion}}
=
\mathrm{MSE}
\left(
\hat{y},
y
\right).
\end{equation}
To encourage every semantic branch to preserve independently discriminative
skill information, the auxiliary stream predictions are also supervised:
\begin{equation}
\mathcal{L}_{\mathrm{stream}}
=
\frac{1}{|\mathcal{K}|}
\sum_{k\in\mathcal{K}}
\mathrm{MSE}
\left(
\hat{y}_{k},
y
\right).
\end{equation}
The complete training objective is
\begin{equation}
\mathcal{L}
=
\mathcal{L}_{\mathrm{fusion}}
+
\mathcal{L}_{\mathrm{stream}},
\label{eq:complete_training_objective}
\end{equation}

\revision{
The fused loss optimizes the final video-level prediction, while the
per-stream auxiliary loss encourages each semantic stream to retain independently
discriminative skill information rather than relying entirely on the global
fusion pathway.
}

\begin{figure*}
    \revisionfloat
    \centering
    \includegraphics[width=1\linewidth]{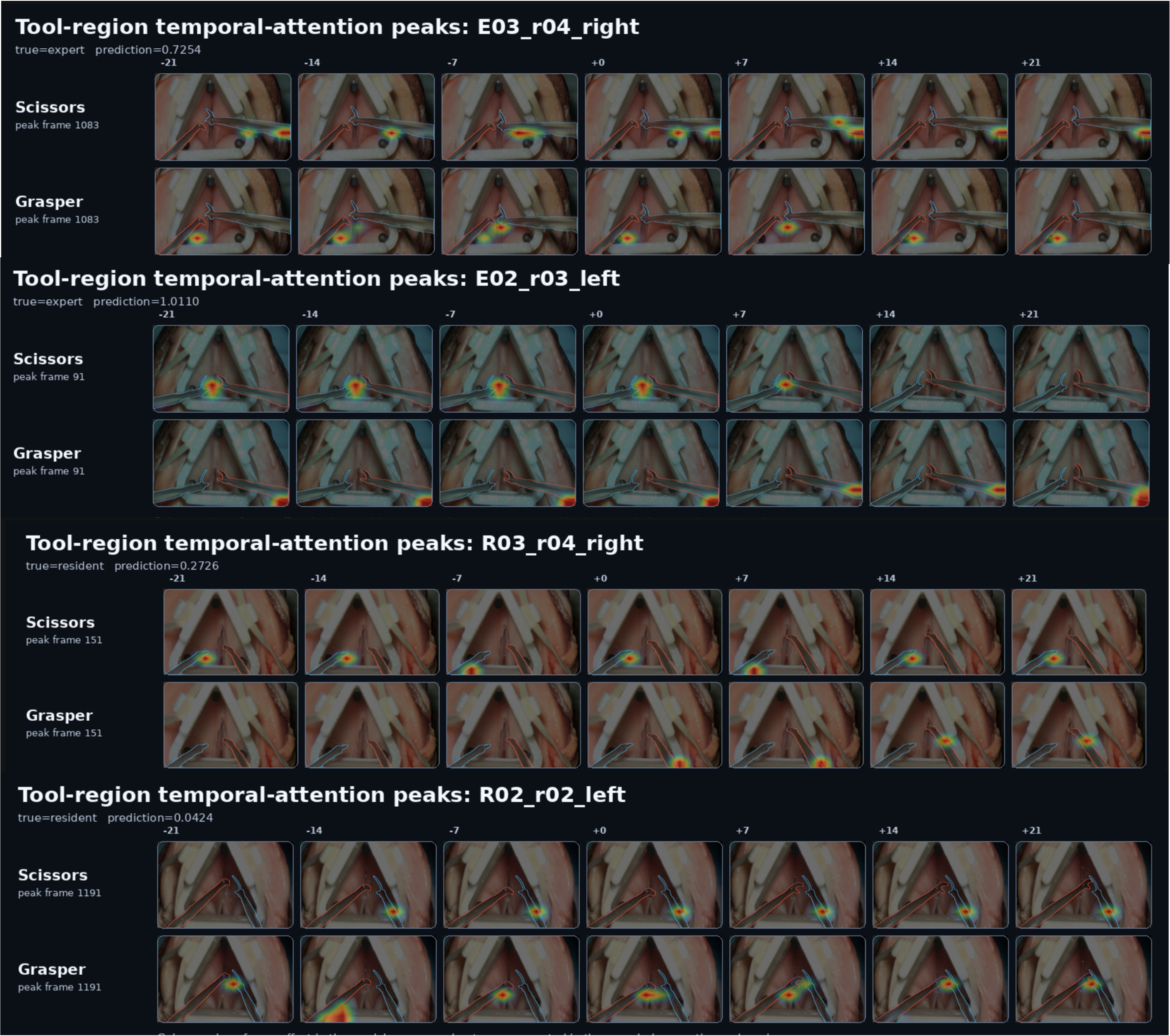}
\caption{\revision{Spatial attention at stream-specific peak-attention moments for four representative trials, including two expert and two resident trials. For each trial, seven frames are sampled around the peak of the combined temporal and semantic-stream attention response, $\alpha_t\gamma_{t,k}$, independently for the scissors and grasper streams. The first row shows the scissors sequence and the second row the grasper sequence, with each sequence centered on its corresponding peak-attention frame. Spatial attention maps are restricted to the respective encoded semantic-mask region and upsampled to the original image resolution. Warmer colors indicate greater spatial attention.}}
    \label{fig:spatial_attention}
\end{figure*}



\section{Experiments and Results}

\paragraph*{Dataset}
\label{sub:dataset}

The dataset contains 33 videos of laryngeal microsurgery recorded using a
stereo surgical microscope. Each video consists of synchronized left- and
right-view microscopic video sequences of the same procedure.
\revision{
The videos had a mean duration of $47.6 \pm 16.5$~s, with a range of
22.8--79.0~s and a total recording duration of 26.16~min.
}

\revision{
The surgical procedures were performed by six surgeons representing two professional training levels: three attending (expert) surgeons and three surgical residents.
The dataset contains 15 procedures performed by the attending surgeons and 18 procedures performed by the residents,
as summarized in
Table~\ref{tab:dataset_summary}. 
}

Two surgical instruments are present in the operative scene: a grasper and
a pair of scissors. Because either instrument may appear on the left or right
side of the image, each video includes tool-identity annotations that map the
left- and right-side instrument regions to the grasper and scissors.
These annotations allow \revision{instrument-specific} representations to be
organized consistently across videos.

\begin{table}[t]
\revisionfloat
\centering
\caption{\revision{Participant composition of the laryngeal microsurgery dataset.}}
\label{tab:dataset_summary}
\small
\begin{tabular}{lcc}
\toprule
Experience level & Participants (videos) & Total videos \\
\midrule
Expert   & E01 (6), E02 (4), E03 (5) & 15 \\
Resident & R01 (6), R02 (6), R03 (6) & 18 \\
\midrule
Overall  & 6 participants & 33 \\
\bottomrule
\end{tabular}
\end{table}

\begin{figure}[t]
    \revisionfloat
    \centering
    \includegraphics[width=\linewidth]{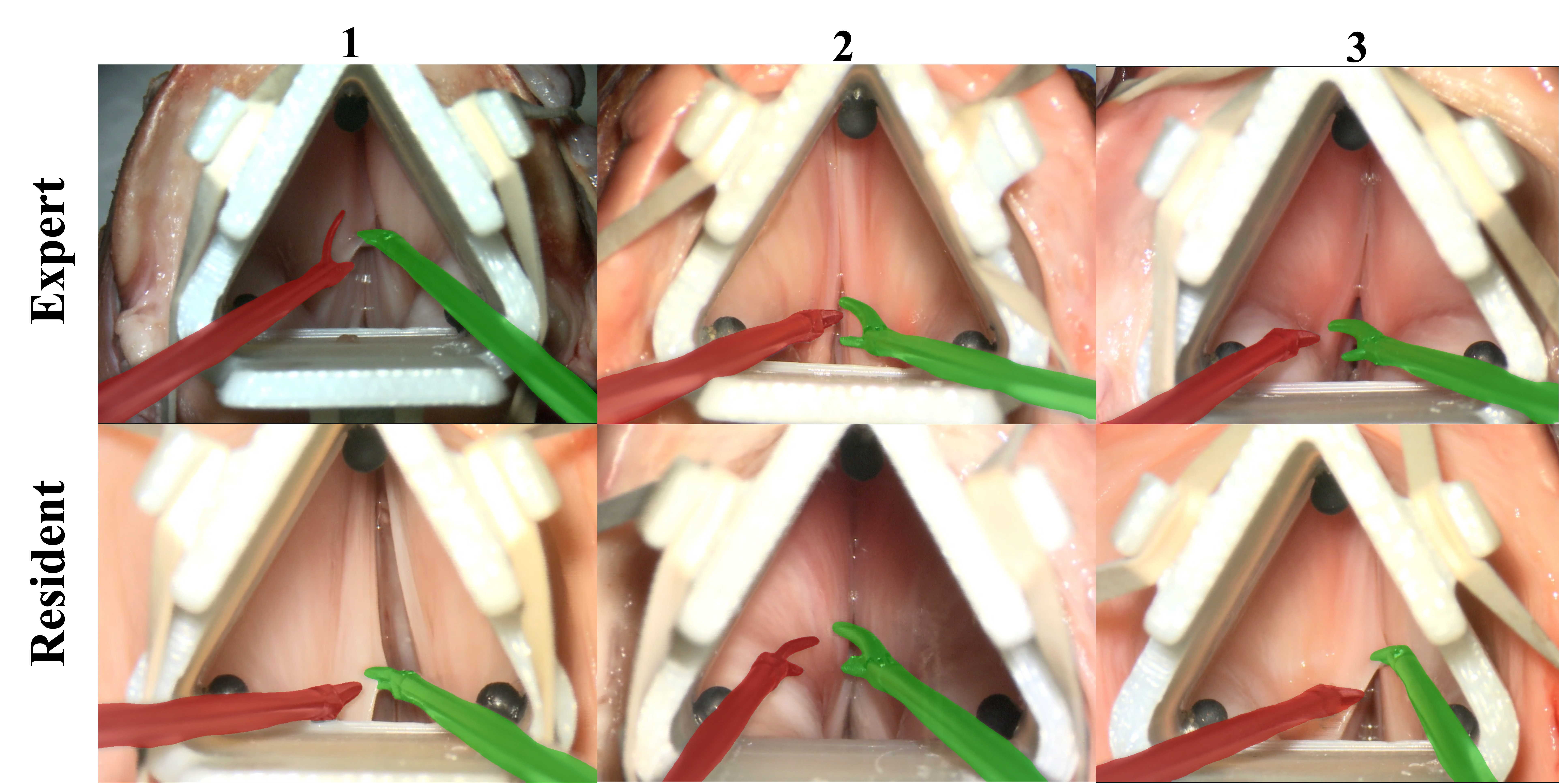}
    \caption{\revision{Representative segmentation results from the six videos used for
    quantitative evaluation. One frame is shown from each participant, including
    three expert and three resident cases, illustrating propagated tool
    segmentation across different users and video sequences.}}
    \label{fig:segmentation_results}
\end{figure}

\revision{
\paragraph*{Segmentation Initialization and Propagation Evaluation}

To evaluate segmentation robustness, we selected one representative video trial
from each participant, resulting in six videos in total (three expert and three
resident). Dice similarity was computed between the segmentation masks and
reference annotations for the manually initialized first frame and across the
full CUTIE-propagated sequence. The mean Dice score was 0.9582 at initialization
and 0.9731 over the propagated sequences, indicating that segmentation accuracy
remained stable following the one-time initialization. Representative results
from all six videos are shown in Fig.~\ref{fig:segmentation_results}, with
quantitative results reported in Table~\ref{tab:segmentation-dice}.
}

\begin{table}[t]
\revisionfloat
\centering
\caption{\revision{Quantitative evaluation of segmentation initialization and propagation
for one video trial from each participant.}}
\label{tab:segmentation-dice}

\small
\setlength{\tabcolsep}{4pt}
\renewcommand{\arraystretch}{0.95}

\begin{tabular}{lcc}
\toprule
\textbf{Trial} &
\textbf{Initialization Dice} &
\textbf{Sequence Dice} \\
\midrule
E01 & 0.9706 & 0.9792 \\
E02 & 0.9594 & 0.9899 \\
E03 & 0.9728 & 0.9513 \\
R01 & 0.9588 & 0.9742 \\
R02 & 0.9343 & 0.9683 \\
R03 & 0.9533 & 0.9759 \\
\midrule
\textbf{Mean} & \textbf{0.9582} & \textbf{0.9731} \\
\bottomrule
\end{tabular}
\end{table}

\revision{
\paragraph*{Task definition, training, and evaluation.}

Using the participant-level training designations, we formulate surgical skill
assessment as a binary classification task \revision{between attending surgeons and surgical residents}. Each video is labeled according to
the training level of the participant, with resident videos
assigned $y=0$ and expert videos assigned $y=1$. The ground-truth label
therefore reflects the participant's training status rather than a video-specific  performance score.

During training, the model learns a scalar video-level output $\hat{y}$ using
the objective in Eq.~\ref{eq:complete_training_objective}. MSE is used as a
regression surrogate for the binary expertise label rather than to estimate a
continuous surgical-skill score. 
For evaluation, we performed leave-one-surgeon-out cross validation.
All six participant-disjoint folds used the
same fixed training settings: AdamW for 20 epochs with learning rate
$2\times10^{-4}$, batch size 1, weight decay $1\times10^{-4}$, and cosine
annealing to $1\times10^{-5}$. Model selection and training settings were kept fixed across all folds, with no inner validation split or fold-specific hyperparameter tuning.

At inference, the scalar output is converted to a binary prediction using a
fixed midpoint threshold. A video score is classified as expert if
$\hat{y} \geq 0.5$ and as resident otherwise. This threshold was specified
a priori and was not tuned on the evaluation data. We report expert-class
precision, recall, and F1-score, together with overall classification
accuracy, under this decision rule.
}

\begin{table}[t]
\revisionfloat
\centering
\caption{\revision{Evaluation results of leave-one-surgeon-out cross-validation. Best values are shown in bold. NB: $^{*}$USSA (RGB) ~\cite{liu2021towards} predicted all trials as resident, resulting in zero precision, recall, and F1 for the expert class.}}
\label{tab:model-comparison}

\scriptsize
\setlength{\tabcolsep}{2.7pt}
\renewcommand{\arraystretch}{0.95}

\begin{tabular}{@{}lrrrrrr@{}}
\toprule
Model & MAE $\downarrow$ & $\rho$ $\uparrow$ & Acc. $\uparrow$ & Prec. $\uparrow$ & Rec. $\uparrow$ & F1 $\uparrow$ \\
\midrule
\multicolumn{7}{@{}l}{\textit{Individual representation baselines}} \\
RGB & 0.267 & \textbf{0.863} & 0.788 & \textbf{1.000} & 0.533 & 0.696 \\
Depth (ASPP) & 0.487 & 0.147 & 0.576 & 0.519 & 0.933 & 0.667 \\
Depth (SPP) & 0.381 & 0.748 & \textbf{0.939} & \textbf{1.000} & 0.867 & 0.929 \\[1pt]
\midrule
\multicolumn{7}{@{}l}{\textit{Proposed model}} \\
\textbf{Depth (SPP) + RGB} & \textbf{0.227} & \textbf{0.863} & \textbf{0.939} & 0.882 & \textbf{1.000} & \textbf{0.938} \\[1pt]
\midrule
\multicolumn{7}{@{}l}{\textit{Architecture ablations}} \\
\quad w/o semantic & 3.184 & -0.275 & 0.303 & 0.300 & 0.400 & 0.343 \\
\quad w/o attention & 1.214 & 0.205 & 0.515 & 0.484 & \textbf{1.000} & 0.652 \\
\quad w/o auxiliary loss & 2.558 & -0.358 & 0.364 & 0.400 & 0.800 & 0.533 \\[1pt]
\midrule
\multicolumn{7}{@{}l}{\textit{Published baseline}} \\[-2pt]
USSA (RGB) & 0.522 & -0.345 & 0.545 & 0.000* & 0.000* & 0.000* \\
\bottomrule
\end{tabular}

\vspace{1pt}
\raggedright
\scriptsize
$\rho$: Spearman correlation; ASPP: atrous spatial pyramid pooling;
SPP: spatial pyramid pooling; USSA: Unified Surgical Skill Assessment.
\end{table}

\begin{table*}[t]
\revisionfloat
\centering
\caption{\revision{Class-specific F1 scores for each held-out surgeon under participant-disjoint cross-validation. $n$ denotes the number of validation trials, and each model column reports F1 for the corresponding expertise class. $^{*}$USSA (RGB)~\cite{liu2021towards} predicted all trials as resident; consequently, F1 is zero for each held-out expert surgeon.}}
\label{tab:per-fold-f1}
\scriptsize
\resizebox{\textwidth}{!}{%
\begin{tabular}{llrrrrrrrrr}
\toprule
Held-out & Class & $n$ & Semantic RGB & Depth (ASPP) & Depth (SPP) & Depth (SPP) + RGB & w/o semantic & w/o attention & w/o auxiliary & USSA (RGB)\\
\midrule
E01 & Expert & 6 & 0.286 & 0.667 & 0.909 & 1.000 & 0.000 & 1.000 & 0.667 & 0.000 \\
E02 & Expert & 4 & 1.000 & 0.800 & 0.857 & 1.000 & 1.000 & 1.000 & 1.000 & 0.000 \\
E03 & Expert & 5 & 0.750 & 0.667 & 1.000 & 1.000 & 0.571 & 1.000 & 1.000 & 0.000 \\
R01 & Resident & 6 & 1.000 & 0.800 & 1.000 & 0.800 & 0.000 & 0.000 & 0.000 & 1.000 \\
R02 & Resident & 6 & 1.000 & 0.571 & 1.000 & 1.000 & 0.800 & 0.000 & 0.000 & 1.000 \\
R03 & Resident & 6 & 1.000 & 0.571 & 1.000 & 1.000 & 0.000 & 0.500 & 0.000 & 1.000 \\
\bottomrule
\end{tabular}%
}
\end{table*}

\revision{
\paragraph*{Experimental Setup and Model Variants}
We evaluated several feature representations to assess the contributions of semantic RGB appearance and depth geometry. The RGB model uses features extracted independently from the grasper, scissors, and anatomy regions. Depth features were encoded using either spatial pyramid pooling (SPP) or atrous spatial pyramid pooling (ASPP), and the proposed multimodal model combines semantic RGB with SPP-based depth features. We additionally ablated semantic decomposition, temporal attention, and auxiliary supervision. All proposed-model variants and ablations were evaluated using the same participant-disjoint leave-one-participant-out protocol. For comparison with prior work, we adapted the publicly available Unified Surgical Skill Assessment (USSA) framework~\cite{liu2021towards} as a full-frame RGB baseline and evaluated it using the same six participant-disjoint folds, trial labels, and pooled out-of-fold metrics, with predictions thresholded at 0.5.
}

\paragraph*{Role of dense 3D scene information}

The primary objective of this work is to investigate whether incorporating dense
metric three-dimensional scene information improves video-based surgical skill
assessment.

\revision{
Table~\ref{tab:model-comparison} compares the performance of the different
semantic-stream representations under the participant-disjoint evaluation
protocol. The results show that the proposed combination of semantic RGB with
SPP-based depth achieves the strongest overall performance, with the lowest MAE
and highest F1 score. SPP-based depth alone also performs strongly, whereas the
ASPP depth representation is less effective, indicating that the benefit of
geometric information depends on how it is encoded.

These results suggest that reconstructed depth provides complementary geometric
information rather than replacing visual appearance. While RGB features encode
instrument and tissue appearance, dense metric depth provides information about
instrument positioning and spatial relationships within the surgical field.
Their combination therefore provides a richer representation of microsurgical
performance than either modality alone.

The architectural ablations in Table~\ref{tab:model-comparison} further
indicate that the complete model benefits from semantic decomposition, temporal
attention, and auxiliary supervision. Removing semantic decomposition resulted
in the largest performance decrease, with the F1 score dropping from 0.938 to
0.343. Removing temporal attention and auxiliary supervision reduced the F1
score to 0.652 and 0.533, respectively. These results suggest that each
component contributes to the overall performance of the framework under the
participant-disjoint evaluation setting.

The participant-wise results in Table~\ref{tab:per-fold-f1} provide additional
detail on performance across the held-out surgeons. In comparison, the adapted
USSA baseline performs poorly on the held-out expert trials and predicts them
predominantly as resident cases under this evaluation setting. This suggests
that full-frame RGB features alone may be less robust to participant-level
variation in this dataset.
}

\begin{table}[t]
\revisionfloat
\centering
\caption{\revision{Average per-frame semantic-component attention. Values are mean $\pm$ standard deviation across validation trials; each trial contributes its within-trial mean equally.}}
\label{tab:semantic_stream_attention}
\small
\setlength{\tabcolsep}{4pt}
\begin{tabular}{lrrr}
\toprule
Run & Grasper & Scissors & Anatomy \\
\midrule
Semantic RGB & 0.46 $\pm$ 0.42 & 0.27 $\pm$ 0.32 & 0.26 $\pm$ 0.31 \\
Semantic Depth (SPP) & 0.44 $\pm$ 0.47 & 0.17 $\pm$ 0.33 & 0.39 $\pm$ 0.45 \\
Semantic Depth (ASPP) & 0.36 $\pm$ 0.43 & 0.28 $\pm$ 0.42 & 0.36 $\pm$ 0.43 \\
Semantic Depth (SPP) + RGB & 0.24 $\pm$ 0.37 & 0.28 $\pm$ 0.36 & 0.47 $\pm$ 0.38 \\
\bottomrule
\end{tabular}
\end{table}

\paragraph*{Role of semantic streams}
\revision{
To better understand how different components of the surgical scene contribute to skill assessment, we examine the learned semantic attention distributions summarized in Table~\ref{tab:semantic_stream_attention}.
Semantic RGB places the greatest attention on the grasper stream, whereas the depth-based representations distribute attention more broadly across instruments and anatomy; in particular, ASPP produces a relatively balanced distribution across the three streams.
The proposed SPP + RGB model, which achieves the strongest overall performance in Table~\ref{tab:model-comparison}, shifts attention toward anatomy ($0.47 \pm 0.38$) and scissors ($0.28 \pm 0.36$), while reducing the grasper contribution ($0.24 \pm 0.37$).
This redistribution is consistent with the nature of laryngeal microsurgery, where skill depends strongly on fine-grained scissors--tissue interactions during precise tissue cutting.
}

\begin{table}[t]
\revisionfloat
\centering
\caption{\revision{Normalized global temporal-attention peak location. Values are mean $\pm$ standard deviation across validation trials, with 0 and 1 denoting the first and final analyzed frames, respectively.}}
\begin{tabular}{llr}
\toprule
Run & Expertise & Peak Location \\
\midrule
Semantic RGB & Expert & 0.47 $\pm$ 0.42 \\
 & Resident & 0.50 $\pm$ 0.29 \\
\midrule
Semantic Depth (SPP) & Expert & 0.49 $\pm$ 0.26 \\
 & Resident & 0.99 $\pm$ 0.01 \\
\midrule
Semantic Depth (ASPP) & Expert & 0.46 $\pm$ 0.40 \\
 & Resident & 0.44 $\pm$ 0.46 \\
\midrule
Semantic Depth (SPP) + RGB & Expert & 0.57 $\pm$ 0.44 \\
 & Resident & 0.59 $\pm$ 0.36 \\
\bottomrule
\label{tab:peak_analysis}
\end{tabular}
\end{table}

\paragraph*{Temporal attention}
\revision{
We further analyzed the global temporal-fusion behavior to determine how
different feature representations guide the model toward different portions of
the procedure. Table~\ref{tab:peak_analysis} shows that the
location of the global temporal-attention peak varies substantially across
representations. Semantic RGB and ASPP depth produce broadly distributed peak
locations for both experts and residents. In contrast, SPP depth exhibits a
strong end-of-sequence preference for resident trials ($0.99 \pm 0.01$), while
expert peaks remain substantially earlier ($0.49 \pm 0.26$). When SPP depth is
combined with Semantic RGB, the resident peak shifts away from the sequence
boundary to $0.59 \pm 0.36$, with a comparable distribution for experts
($0.57 \pm 0.44$). This suggests that multimodal fusion mitigates the pronounced
terminal attention bias observed with SPP depth alone and yields a more balanced
temporal representation across expertise groups.
}

\begin{figure}[h]
    \revisionfloat
    \centering
    \includegraphics[width=1\linewidth]{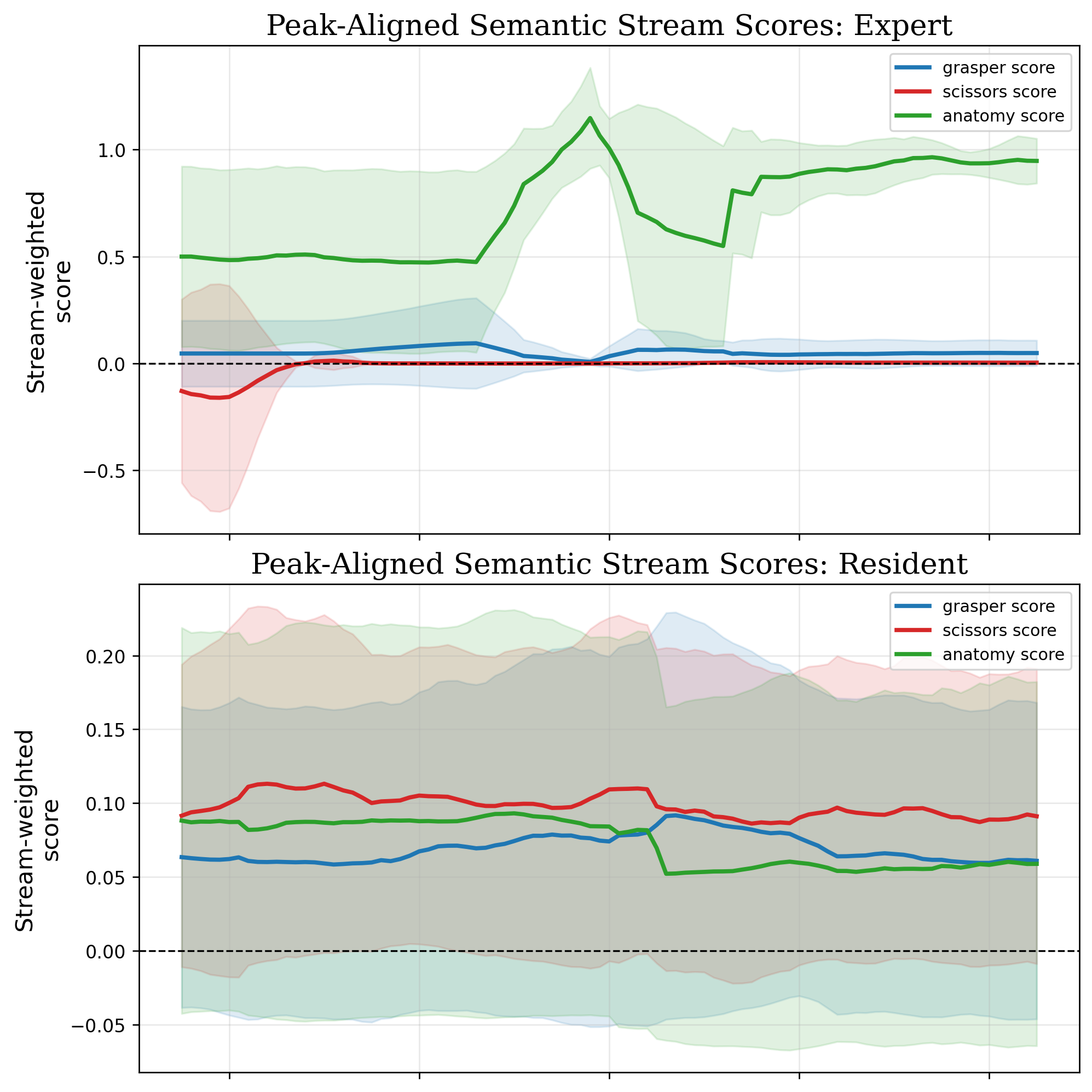}
    \caption{\revision{Stream-weighted local scores aligned to the peak temporal-attention frame. The horizontal axis represents time relative to the frame receiving the maximum global temporal-attention weight. Solid lines show the mean across trials, and shaded regions indicate the standard deviation.}}
    \label{fig:peak_aligned_scores}
\end{figure}

\revision{
\paragraph*{Semantic and spatial attention}
To further examine how semantic information contributes around highly attended
temporal regions, Fig.~\ref{fig:peak_aligned_scores} shows the stream-weighted
local scores aligned to the peak temporal-attention frame. Expert trials exhibit
a pronounced anatomy contribution around the attention peak, while the grasper
and scissors contributions remain comparatively small. In contrast, resident
trials show lower-magnitude and more evenly distributed contributions across
the three semantic streams, with the scissors stream contributing modestly more
than the grasper and anatomy streams. These patterns suggest that highly attended
regions in expert trials are more strongly associated with anatomical context,
whereas resident predictions rely on a more distributed combination of semantic
features.

We further examine the spatial distribution of attention in Fig.~\ref{fig:spatial_attention}, which presents four representative trials, including two expert and two resident examples. 
An attending otolaryngologist who did not participate in the recorded trials independently reviewed these representative attention maps and identified the instrument tip and instrument pivoting region as prominent areas of model attention. The attending surgeon considered these spatial regions clinically plausible in the context of instrument–tissue interaction and instrument stabilization by pivoting it on the laryngoscope, which are both relevant for surgical task execution.
}

\section{Conclusions}
\label{sec:Conclusions}

This work presented a semantic RGB--depth approach for automated surgical
skill assessment from microscopic stereo videos. 
The proposed approach incorporates dense depth representation of the surgical scene by combining sparse stereo depth with
dense monocular depth estimates through a regression-based depth-fusion
strategy. The reconstructed geometric representation is integrated with
semantically decomposed RGB streams using a hierarchical attention
architecture that jointly models instrument appearance, scene geometry,
instrument motion, and instrument--tissue interactions.

\revision{
Evaluation on a dataset comprising of \textit{exvivo} transoral microlaryngeal procedures using leave-one-surgeon-out validation
showed that depth-derived geometric information provides complementary information for
skill assessment beyond RGB appearance alone. The semantic representation
further enables qualitative examination of spatial, temporal, and semantic
attention patterns. An expert surgeon considered the observed spatial
attention patterns clinically plausible, noting greater scissors attention
near tool--tissue interaction regions and greater grasper attention near
pivoting regions. These attention patterns are interpreted as qualitative
visualizations of learned model weighting; we acknowledge that causality
between the highlighted regions and the predicted skill level cannot be
directly established from attention weights alone.
}

\revision{
Although this study focused on \textit{exvivo} transoral microlaryngeal surgery from
a single dataset, the proposed framework may provide a basis for extending
RGB--depth skill assessment to other microscope-guided procedures in which
stereo video is available. A limitation of the present study is that the
reconstructed depth was not evaluated against dense per-pixel metric ground
truth. Future work will investigate validation using more comprehensive
geometric ground truth, as well as larger, multi-institutional and \textit{invivo}
datasets. We will also examine whether highly attended regions exert greater
influence on model predictions under controlled perturbations. Additional
directions include self-supervised learning from unlabeled surgical videos,
continuous skill scoring, and real-time intraoperative feedback.
}

\bibliography{wileyNJD-MLA}

@inproceedings{liu2021towards,
  title={Towards unified surgical skill assessment},
  author={Liu, Daochang and Li, Qiyue and Jiang, Tingting and Wang, Yizhou and Miao, Rulin and Shan, Fei and Li, Ziyu},
  booktitle={Proceedings of the IEEE/CVF Conference on Computer Vision and Pattern Recognition},
  pages={9522--9531},
  year={2021}
}

@inproceedings{li2022surgical,
  title={Surgical skill assessment via video semantic aggregation},
  author={Li, Zhenqiang and Gu, Lin and Wang, Weimin and Nakamura, Ryosuke and Sato, Yoichi},
  booktitle={International Conference on Medical Image Computing and Computer-Assisted Intervention},
  pages={410--420},
  year={2022},
  organization={Springer}
}

@article{anastasiou2023keep,
  title={Keep Your Eye on the Best: Contrastive Regression Transformer for Skill Assessment in Robotic Surgery},
  author={Anastasiou, Dimitrios and Jin, Yueming and Stoyanov, Danail and Mazomenos, Evangelos},
  journal={IEEE Robotics and Automation Letters},
  volume={8},
  number={3},
  pages={1755--1762},
  year={2023},
  publisher={IEEE}
}

@inproceedings{ding2023sedskill,
  title={SEDSkill: Surgical Events Driven Method for Skill Assessment from Thoracoscopic Surgical Videos},
  author={Ding, Xinpeng and Xu, Xiaowei and Li, Xiaomeng},
  booktitle={International Conference on Medical Image Computing and Computer-Assisted Intervention},
  pages={35--45},
  year={2023},
  organization={Springer}
}

@article{zia2018automated,
  title={Automated surgical skill assessment in RMIS training},
  author={Zia, Aneeq and Essa, Irfan},
  journal={International journal of computer assisted radiology and surgery},
  volume={13},
  pages={731--739},
  year={2018},
  publisher={Springer}
}

@article{wisotzky2025intraoperative,
  title={Intraoperative measurements in stapedotomy using 3D stereo imaging for optimal prosthesis length selection},
  author={Wisotzky, Eric L and Lausch, Felix and Kienle, Linus L and van Bonn-Ytrehus, Sara and Hilsmann, Anna and Eisert, Peter and Schraven, Sebastian P and Mlynski, Robert},
  journal={Computers in biology and medicine},
  volume={191},
  pages={110233},
  year={2025},
  publisher={Elsevier}
}

@article{ha2024target,
  title={Target-specified reference-based deep learning network for joint image deblurring and resolution enhancement in surgical zoom lens camera calibration},
  author={Ha, Ho-Gun and Jeung, Deokgi and Ullah, Ihsan and Tokuda, Junichi and Hong, Jaesung and Lee, Hyunki},
  journal={Computers in Biology and Medicine},
  volume={183},
  pages={109309},
  year={2024},
  publisher={Elsevier}
}

@article{funke2019video,
  author  = {Funke, Isabel and Mees, S{\"o}ren Torge and Weitz, J{\"u}rgen and Speidel, Stefanie},
  title   = {Video-based surgical skill assessment using 3D convolutional neural networks},
  journal = {International Journal of Computer Assisted Radiology and Surgery},
  year    = {2019},
  volume  = {14},
  number  = {7},
  pages   = {1217--1225},
  doi     = {10.1007/s11548-019-01995-1},
  pmid    = {31104257}
}

@article{wan2024spatiotemporal,
  author  = {Wan, Bohua and Peven, Michael and Hager, Gregory D. and Sikder, Shameema and Vedula, S. Swaroop},
  title   = {Spatial-temporal attention for video-based assessment of intraoperative surgical skill},
  journal = {Scientific Reports},
  year    = {2024},
  volume  = {14},
  pages   = {26912},
  doi     = {10.1038/s41598-024-77176-1}
}

@article{hira2022video,
  author  = {Hira, Sanchit and Singh, Digvijay and Kim, Tae Soo and Gupta, Shobhit and Hager, Gregory and Sikder, Shameema and Vedula, S. Swaroop},
  title   = {Video-Based Assessment of Intraoperative Surgical Skill},
  journal = {International Journal of Computer Assisted Radiology and Surgery},
  year    = {2022},
  volume  = {17},
  number  = {10},
  pages   = {1801--1811},
  doi     = {10.1007/s11548-022-02681-5}
}

@inproceedings{yang2024depthanythingv2,
  author    = {Yang, Lihe and Kang, Bingyi and Huang, Zilong and Zhao, Zhen and Xu, Xiaogang and Feng, Jiashi and Zhao, Hengshuang},
  title     = {Depth Anything V2},
  booktitle = {Advances in Neural Information Processing Systems},
  year      = {2024}
}

@inproceedings{kirillov2023segment,
  title={Segment Anything},
  author={Kirillov, Alexander and Mintun, Eric and Ravi, Nikhila and Mao, Hanzi and Rolland, Chloe and Gustafson, Laura and Xiao, Tete and Whitehead, Spencer and Berg, Alexander C. and Lo, Wan-Yen and Doll{\'a}r, Piotr and Girshick, Ross},
  booktitle={Proceedings of the IEEE/CVF International Conference on Computer Vision},
  pages={4015--4026},
  year={2023}
}

@inproceedings{cheng2024putting,
  title={Putting the Object Back into Video Object Segmentation},
  author={Cheng, Ho Kei and Oh, Seoung Wug and Price, Brian and Lee, Joon-Young and Schwing, Alexander},
  booktitle={Proceedings of the IEEE/CVF Conference on Computer Vision and Pattern Recognition},
  pages={3151--3161},
  year={2024}
}

@misc{opencv_stereobm,
  author       = {{OpenCV}},
  title        = {{cv::StereoBM Class Reference}},
  howpublished = {\url{https://docs.opencv.org/3.4/d9/dba/classcv_1_1StereoBM.html}},
  year         = {2024},
  note         = {Accessed: 2026-06-18}
}

@article{lavanchy2021automation,
  title={Automation of surgical skill assessment using a three-stage machine learning algorithm},
  author={Lavanchy, Jo{\"e}l L and Zindel, Joel and Kirtac, Kadir and Twick, Isabell and Hosgor, Enes and Candinas, Daniel and Beldi, Guido},
  journal={Scientific reports},
  volume={11},
  number={1},
  pages={5197},
  year={2021},
  publisher={Nature Publishing Group UK London}
}

@article{cho2024tremor,
  title={Tremor assessment in robot-assisted microlaryngeal surgery using computer vision-based tool tracking},
  author={Cho, Sue M and Joo, Henry H and Golla, Pranathi and Sahu, Manish and Shankar, Ahjeetha and Trakimas, Danielle R and Creighton, Francis and Akst, Lee and Taylor, Russell H and Galaiya, Deepa},
  journal={Otolaryngology--Head and Neck Surgery},
  volume={171},
  number={1},
  pages={188--196},
  year={2024},
  publisher={Wiley Online Library}
}

@article{martin1997objective,
  title={Objective structured assessment of technical skill (OSATS) for surgical residents},
  author={Martin, JA and Regehr, Glenn and Reznick, Richard and Macrae, Helen and Murnaghan, John and Hutchison, Carol and Brown, M},
  journal={British journal of surgery},
  volume={84},
  number={2},
  pages={273--278},
  year={1997},
  publisher={Wiley Online Library}
}

@article{niitsu2013Objective,
  title={Using the Objective Structured Assessment of Technical Skills (OSATS) global rating scale to evaluate the skills of surgical trainees in the operating room},
  author={Niitsu, Hiroaki and Hirabayashi, Naoki and Yoshimitsu, Masanori and Mimura, Takeshi and Taomoto, Junya and Sugiyama, Yoich and Murakami, Shigeru and Saeki, Shuji and Mukaida, Hidenori and Takiyama, Wataru},
  journal={Surgery today},
  volume={43},
  number={3},
  pages={271--275},
  year={2013},
  publisher={Springer}
}

@article{reznick2006teaching,
  title={Teaching surgical skills—changes in the wind},
  author={Reznick, Richard K and MacRae, Helen},
  journal={New England Journal of Medicine},
  volume={355},
  number={25},
  pages={2664--2669},
  year={2006},
  publisher={Mass Medical Soc}
}

@inproceedings{wang2020towards,
  title={Towards accurate and interpretable surgical skill assessment: A video-based method incorporating recognized surgical gestures and skill levels},
  author={Wang, Tianyu and Wang, Yijie and Li, Mian},
  booktitle={International Conference on Medical Image Computing and Computer-Assisted Intervention},
  pages={668--678},
  year={2020},
  organization={Springer}
}

@article{kelly2020bidirectional,
  title={Bidirectional long short-term memory for surgical skill classification of temporally segmented tasks},
  author={Kelly, Jason D and Petersen, Ashley and Lendvay, Thomas S and Kowalewski, Timothy M},
  journal={International journal of computer assisted radiology and surgery},
  volume={15},
  number={12},
  pages={2079--2088},
  year={2020},
  publisher={Springer}
}

@article{wang2018deep,
  title={Deep learning with convolutional neural network for objective skill evaluation in robot-assisted surgery},
  author={Wang, Ziheng and Majewicz Fey, Ann},
  journal={International journal of computer assisted radiology and surgery},
  volume={13},
  number={12},
  pages={1959--1970},
  year={2018},
  publisher={Springer}
}

@article{zia2018video,
  title={Video and accelerometer-based motion analysis for automated surgical skills assessment},
  author={Zia, Aneeq and Sharma, Yachna and Bettadapura, Vinay and Sarin, Eric L and Essa, Irfan},
  journal={International journal of computer assisted radiology and surgery},
  volume={13},
  number={3},
  pages={443--455},
  year={2018},
  publisher={Springer}
}
\clearpage
\onecolumn

\end{document}